\documentclass[10pt,twocolumn,letterpaper]{article}

\usepackage{iccv}              
\usepackage{multirow}
\usepackage{booktabs}
\usepackage{tabularx}
\usepackage{xcolor}         
\usepackage{algorithm}
\usepackage{algorithmic}
\usepackage{epsfig}
\usepackage{graphicx}
\usepackage{amsmath}
\usepackage{amssymb}
\usepackage{bm}
\usepackage{arydshln}
\usepackage{wrapfig}
\usepackage{pifont}
\usepackage{newunicodechar}
\newunicodechar{✓}{\ding{51}}

\definecolor{iccvblue}{rgb}{0.21,0.49,0.74}
\definecolor{lightyellow}{rgb}{0.9, 0.85, 0.55}
\usepackage[pagebackref,breaklinks,colorlinks,allcolors=iccvblue]{hyperref}

\def\paperID{3275} 
\def\confName{ICCV}
\def\confYear{2025}

\title{Uncertainty-Aware Gradient Stabilization for Small Object Detection}

\author{
\textbf{Huixin Sun$^{1}$, Yanjing Li$^{1}$,} 
\textbf{Linlin Yang$^{4}$, Xianbin Cao$^{1\ast}$, Baochang Zhang$^{2,3}$\thanks{Corresponding authors. Email: {\tt\small sunhuixin@buaa.edu.cn}.}} \\
$^{1}$School of Electronic Information Engineering, Beihang University \\
$^{2}$School of Artificial Intelligence, Beihang University \\
$^{3}$Zhongguancun Laboratory, Beijing, China \\
$^{4}$State Key Laboratory of Media Convergence and Communication, CUC \\
}

\begin{document}
\maketitle
\begin{abstract}
Despite advances in generic object detection, there remains a performance gap in detecting small objects compared to normal-scale objects.  
We reveal that conventional object localization methods suffer from gradient instability in small objects due to sharper loss curvature, leading to a convergence challenge.
To address the issue, we propose Uncertainty-Aware Gradient Stabilization (UGS), a framework that reformulates object localization as a classification task to stabilize gradients.  
UGS quantizes continuous labels into interval non-uniform discrete representations. 
Under a classification-based objective, the localization branch generates bounded and confidence-driven gradients, mitigating instability.
Furthermore, UGS integrates an uncertainty minimization (UM) loss that reduces prediction variance and an uncertainty-guided refinement (UR) module that identifies and refines high-uncertainty regions via perturbations.
Evaluated on four benchmarks, UGS consistently improves anchor-based, anchor-free, and leading small object detectors. 
Especially, UGS enhances DINO-5scale by 2.6 AP on VisDrone, surpassing previous state-of-the-art results.
\end{abstract}    
\section{Introduction}
\label{sec:intro}
Recent advances in deep neural networks (DNNs)~\cite{he2015convolutional,zhang2022dino} have substantially improved the object detection field~\cite{Everingham10,lin2014microsoft}.
Nevertheless, detecting small objects remains a persistent challenge~\cite{yang2022querydet}.
For instance, Cascade R-CNN~\cite{cai2018cascade}, one of the most representative two-stage object detectors, achieves 45.5\% and 55.2\% AP on medium and large-sized objects but only 23.7\% AP on small objects in the COCO test-dev set~\cite{lin2014microsoft}. The significant performance gap between small and normal-sized objects limits the effectiveness of object detectors in real-world applications, such as driving assistance, traffic management, and anomaly detection.

\begin{figure}[t]
    \centering
    \includegraphics[width=\linewidth]{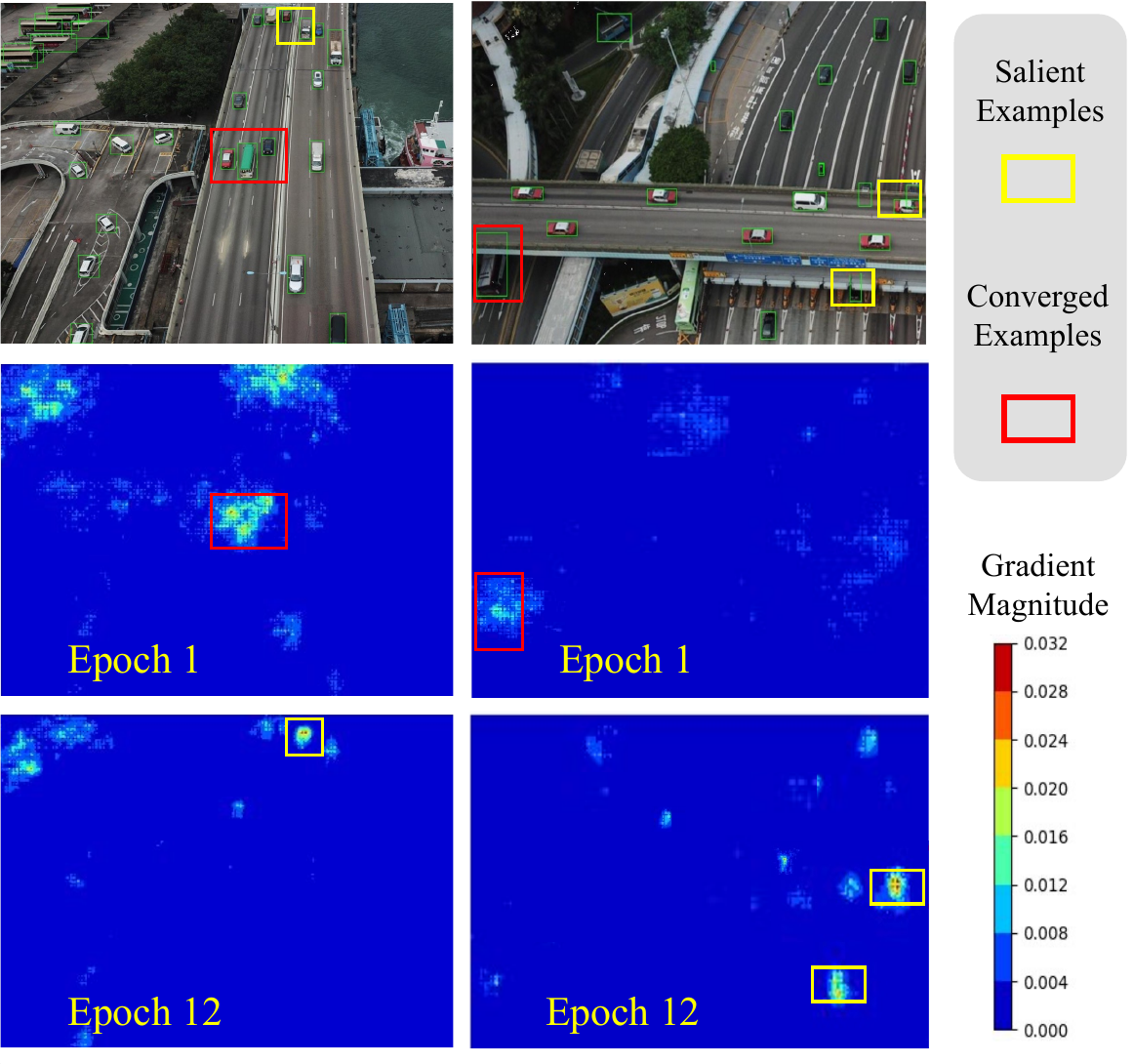}
    \caption{Gradient magnitude maps based on the $\ell_2$-Norm of the gradient in the regression features~\cite{guo2021distilling}. The maps are generated using a 1$\times$ Faster R-CNN~\cite{ren2015faster} detector with ResNet-18 backbone. By epoch 12, most objects exhibit converged gradients (\textcolor{red}{red} areas) while multiple small object examples (\textcolor{lightyellow}{yellow} areas) exhibit salient gradients, indicating a convergence challenge.}
    \label{fig:motivation}
    \vspace{-4mm}
\end{figure}

Due to the limited pixel inputs (under 32×32~\cite{lin2014microsoft}),
one of the primary challenges in small object detection is the extraction of discriminative foreground features~\cite{liang2019small}.
The challenge is compounded in cluttered environments~\cite{cao2024visible}, where occlusions, background noise, and low signal-to-noise ratio conditions induce feature ambiguity.
Consequently, generic detectors can develop a feature bias towards distinguishing the foreground from background regions that resemble it~\cite{xulearning}.
Recent efforts address these problems by increasing the resolution of feature maps~\cite{lin2017feature,akyon2022slicing}, integrating contextual information~\cite{du2023adaptive,yang2024pinwheel}, and employing auxiliary self-reconstruction branches~\cite{cao2024visible,xulearning} to enhance object features.
In this work, we analyze the SOD challenge from a novel perspective of gradient stability.

\begin{figure}[t]
    \centering
    \includegraphics[width=\linewidth]{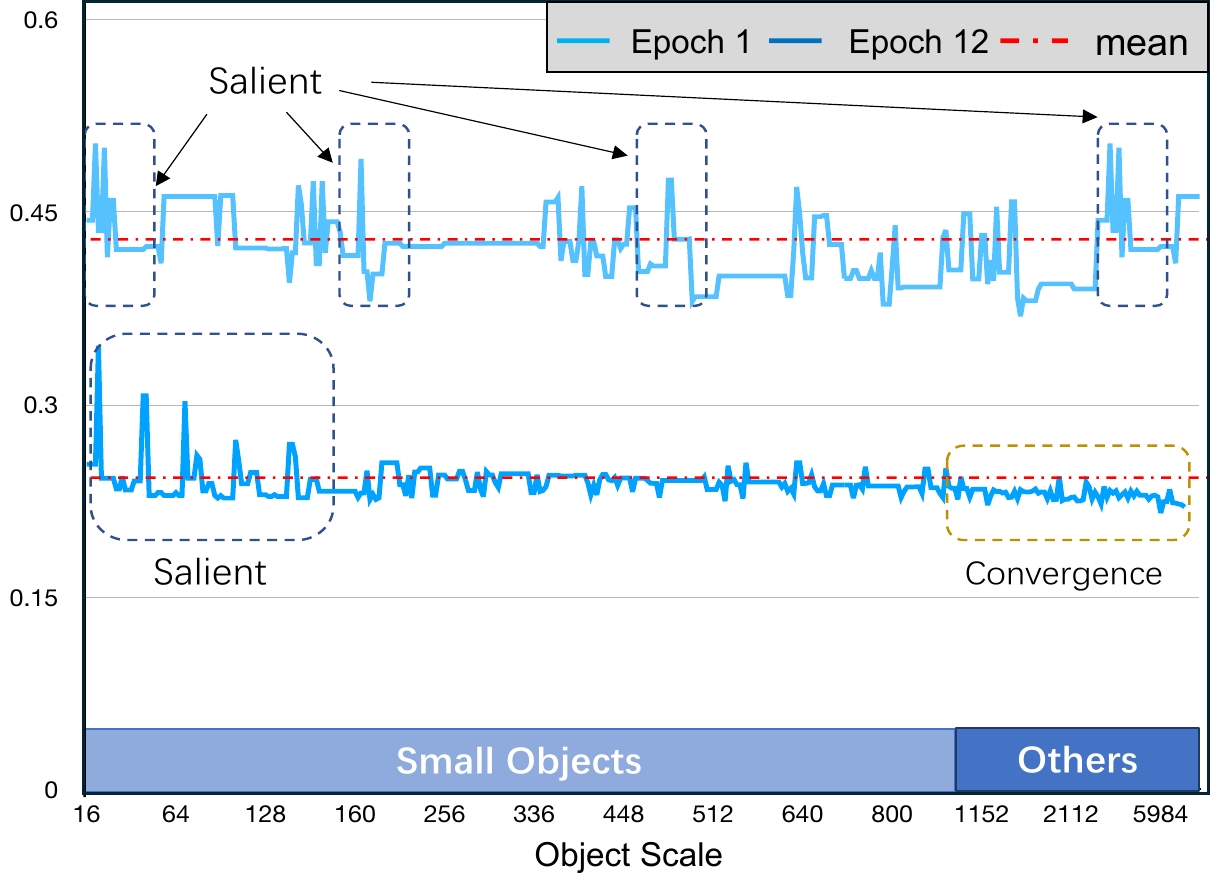}
    \caption{Statistics of the average object gradient. Object gradient is derived by summing the gradient magnitudes within each bounding box based on the gradient maps. The average gradient of each object scale is calculated as the mean across objects in the corresponding size category. The figure is generated using 100 randomly selected samples in the VisDrone {\tt test}~\cite{zhu2018visdrone}. The bottom curve (epoch 12) reveals that gradients for medium and large objects have significantly reduced after training. In contrast, gradients for small objects remain pronounced, indicating the convergence challenge.}
    \label{fig:average_gradient}
    \vspace{-4mm}
\end{figure}

Motivated by prior research on gradient anomalies~\cite{ming2023deep,tan2021equalization}, we examine the Hessian matrix of the object localization losses and demonstrate that conventional methods~\cite{ren2015faster,yu2016unitbox} suffer from gradient instability in small objects due to the sharper loss curvature, which leads to a convergence challenge.
We illustrate this phenomenon using the gradient magnitudes generated from the localization features in Fig.~\ref{fig:motivation}.
As shown, the detector converges well for most objects (\textcolor{red}{red} areas) but exhibits salient gradients in multiple small object examples at epoch 12 (\textcolor{lightyellow}{yellow} areas).
We further quantified the average object gradient based on the gradient magnitudes. 
As shown in Fig.~\ref{fig:average_gradient}, the gradients of small objects remain pronounced at epoch 12, while those of normal-scale objects converge effectively.
%

In light of the analysis, we propose Uncertainty-Aware Gradient Stabilization (UGS), a framework that stabilizes gradients by reformulating regression as a classification task with uncertainty-aware mechanisms.  
The framework of UGS is illustrated in Fig.\,\ref{architecture}. 
Building on prior classification-based localization methods~\cite{qiu2020offset,li2020generalized}, UGS quantizes continuous labels into interval non-uniform discrete representations to address the label imbalance in small objects. 
Under a classification-based objective, the localization branch generates bounded and confidence-driven gradients, mitigating instability.
To further stabilize training, UGS introduces a dual uncertainty-aware mechanism consisting of an {uncertainty minimization (UM)} loss and an {uncertainty-guided refinement (UR)} module.
The UM loss explicitly models and minimizes prediction uncertainty via entropy, reducing the prediction variance.
Additionally, the UR module leverages adversarial perturbations to identify and refine high-uncertainty regions, thereby improving feature robustness.

To summarize, our main contributions are three-fold:
\begin{enumerate}
\item A gradient analysis is conducted to investigate the Small Object Detection (SOD) challenge, which shows that conventional object localization suffers from unstable gradients on small objects and leads to a convergence challenge. We propose a novel Uncertainty-Aware Gradient Stabilization (UGS) method to improve gradient stability and foster better convergence.
\item UGS integrates three key components: a classification-based localization objective for generating bounded and confidence-driven gradients, an uncertainty minimization loss that explicitly models and minimizes prediction uncertainty in small objects, and an uncertainty-guided refinement module that leverages adversarial perturbations to identify and refine high-uncertainty regions.
\item UGS exhibits consistent increases on baseline detectors and state-of-the-art small-object detectors, demonstrating effectiveness in general object detection and high-resolution detection.
\end{enumerate} 

\vspace{-1mm}
\section{Related Work}
\subsection{Small Object Detection} Small object detection poses significant challenges in computer vision, due to the inherent limitations of pixel input~\cite{xulearning}. The main difficulties in small object detection include insufficient feature representation~\cite{lin2017feature,liang2019small}, feature information loss during down-sampling~\cite{kisantal2019augmentation}, and fewer positive samples assigned because of increased sensitivity in IoU calculations~\cite{xu2021dot,xu2022rfla,yuan2023small}. To address these issues, various approaches have been proposed, which can be categorized into four main strategies:
{feature enhancement}, where methods like~\cite{chen2017r,lin2017feature,chen2017deeplab} improve feature representation through advanced architectures and fusion techniques;
{data augmentation and oversampling}, where techniques like~\cite{zoph2020learning,liu2016ssd} increase the diversity and quantity of training data for small objects; 
{scale-aware training}, where approaches like~\cite{li2019scale,singh2018analysis} adapt models to handle objects at multiple scales;
{super-solution-based method}, where frameworks like~\cite{bashir2021small,li2017perceptual} enhance small object features by reconstructing high-resolution representations.
Recent state-of-the-art methods focus on label assignment and proposal refinement to ensure both the quantity and quality of proposals for small objects~\cite{xu2022rfla,yuan2023small}. Additionally, auxiliary self-reconstruction~\cite{cao2024visible, xulearning} and spectral enhancement~\cite{sun2025set} methods have been proposed to enhance the weak representations of objects. 
Our method is orthogonal to existing small object detection methods from a new perspective of solving the convergence challenge in localization. 
\subsection{Uncertainty Estimation in Object Detection}
Recent advances in localization uncertainty estimation aim to quantify prediction uncertainty by modeling bounding box distributions. Early work by ~\cite{jiang2018acquisition} directly predicts the Intersection-over-Union (IoU) between ground-truth and predicted boxes as a confidence measure. Subsequent approaches like ~\cite{he2019bounding} formulate uncertainty via KL divergence loss between Dirac delta ground-truths and Gaussian predictions, enabling variance-based refinement. The state-of-the-art GFL V1~\cite{li2020generalized} further generalizes this paradigm by modeling arbitrary distributions through distribution focal loss.  
However, the methods are based on anchor-free frameworks (e.g., FCOS \cite{tian2019fcos}), limiting their compatibility with anchor-based methods (e.g., RetinaNet~\cite{zhang2020bridging}), modern anchor-free methods (e.g., YOLO series~\cite{redmon2018yolov3, bochkovskiy2020yolov4, zhu2021tph}), two-stage detectors (e.g., Faster R-CNN~\cite{ren2015faster}), and DETR-based detectors (e.g., DINO~\cite{zhang2022dino}).
%
Moreover, while uncertainty estimation addresses challenges such as occlusion and noise~\cite{lee2022localization}, it remains unexplored for small object detection.

\begin{figure}[t]
\begin{center}
\centering
    \centerline{\includegraphics[width=0.9\linewidth]{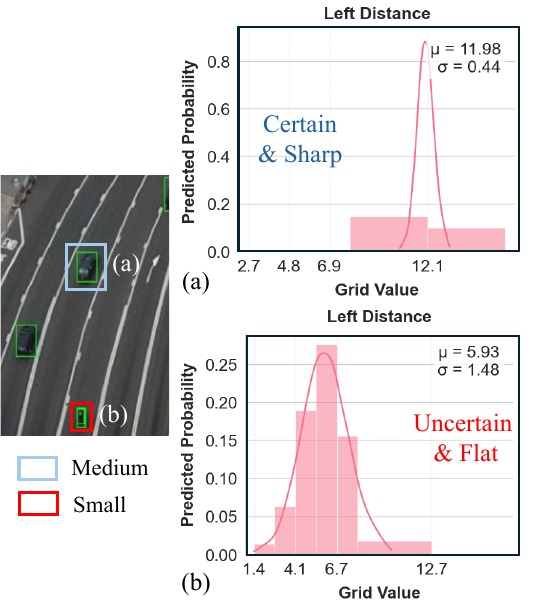}}
    \vspace{-2mm}
    \caption{A qualitative example of localization uncertainty using FCOS~\cite{tian2019fcos} trained with $\mathcal{L}_\text{CE}$ (Eq.\,(\ref{eq:ce})). (a) The medium object instance shows certain and sharp prediction distribution ($\sigma=0.44$), while (b) the small object instance exhibits flatter distribution ($\sigma=1.48$), reflecting higher prediction uncertainty. }
    \label{fig:qualitative_case}
\end{center}
\vspace{-6mm}
\end{figure}
\vspace{-2mm}
\section{Method}
In this section, we first analyze the gradient instability in norm-based and IoU-based localization methods, then introduce the Uncertainty-Aware Gradient Stabilization (UGS) design as a solution.
\begin{figure*}[t]
    \centering
    \includegraphics[width=\linewidth]{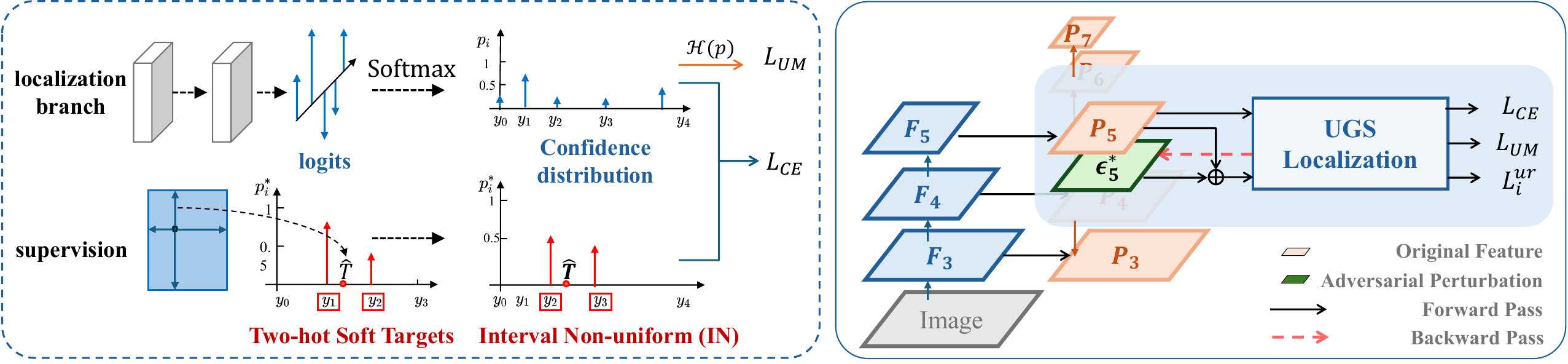}
    \caption{Overview of the proposed UGS localization method with FCOS~\cite{tian2019fcos} framework, consisting a classification-based localization objective ($\mathcal{L}_\text{CE}$), an uncertainty minimization loss ($\mathcal{L}_\text{UM}$), and an uncertainty-guided refinement module ($\mathcal{L}^\text{ur}$).}
    \label{architecture}
\end{figure*}

\subsection{Gradient Instability in Small Objects}  
\label{sec:gradient_instability}  

\noindent{\bf Norm-based Localization.}  
We analyze gradient instability using the $\mathcal{L}_2$ loss in norm-based localization. Following previous detectors~\cite{girshick2014rich,girshick2015fast,ren2015faster}, we denote the localization targets and predictions as:  
\begin{equation}
\begin{split}
    \{T_{x}, T_{y}, T_{w}, T_{h}\} &= \{\frac{x - x_a}{w_a}, \frac{y - y_a}{h_a}, \log\frac{w}{w_a}, \log\frac{h}{h_a}\}, \\
    \{\hat{T}_{x}, \hat{T}_{y}, \hat{T}_{w}, \hat{T}_{h}\} &= \{\frac{\hat{x} - x_a}{w_a}, \frac{\hat{y} - y_a}{h_a}, \log\frac{\hat{w}}{w_a}, \log\frac{\hat{h}}{h_a}\},
\end{split}
\end{equation}
where \((x_a, y_a, w_a, h_a)\) denote the anchor coordinates, \((x, y, w, h)\) the ground-truth coordinates, and \((\hat{x}, \hat{y}, \hat{w}, \hat{h})\) the predicted coordinates, respectively. 
The $\mathcal{L}_2$ loss can be formulated as:
\begin{equation}
    \begin{split}
        \mathcal{L}_{2}(T_{x}, \hat{T}_{x}) = \|T_{x} - \hat{T}_{x}\|_2^2,
    \end{split}
\label{eq1}
\end{equation}
which applies to \(y\), \(w\), and \(h\).
For center coordinates $(x, y)$, the Hessian derives as:
\begin{equation}
    \mathbf{H}_x = \frac{\partial^2 \mathcal{L}_2}{\partial \hat{x}^2} = \frac{2}{w_a^2}, \quad \mathbf{H}_y = \frac{\partial^2 \mathcal{L}_2}{\partial \hat{y}^2} = \frac{2}{h_a^2},
\label{eq:hessian_xy}
\end{equation}
where the Lipschitz constants $K_x$ and $K_y$ scale inversely with the square of anchor size. For small objects assigned with smaller anchors, the Lipschitz constants increase, leading to steeper loss curvature.
For size regression, the Hessian is:
\begin{equation}
    \mathbf{H}_w = \frac{\partial^2 \mathcal{L}_2}{\partial \hat{w}^2} = 2 \cdot \left(\frac{1}{\hat{w}}\right)^2,
\label{eq:hessian_wh}
\end{equation}
where the Lipschitz constant \( K_w \) depends inversely on the predicted width $\hat{w}$, resulting in steeper loss curvature for small objects.
Steep loss curvature can result in unstable updates~\cite{nesterov2018lectures} that are prone to oscillate or diverge near minima, causing a convergence challenge.
As shown in Fig.~\ref{fig:average_gradient}, the gradients of small objects remain pronounced at epoch 12, while those of normal-scale objects converge effectively.

\noindent{\bf IoU-based Localization.}
We analyze the issue using the IoU loss~\cite{yu2016unitbox}, which is applicable to other IoU-based losses~\cite{rezatofighi2019generalized,zheng2020distance}.
Following~\cite{xu2021dot}, we consider the case where the ground truth \((x, y, w, h)\) and the prediction \((\hat{x}, \hat{y}, \hat{w}, \hat{h})\) are aligned square boxes with a center displacement \(d = |x - \hat{x}|\). The IoU loss is expressed as:
\begin{equation}  
\mathcal{L}_{\text{IoU}} = -\ln\left(\frac{I}{U}\right), 
\label{eq6}
\end{equation}  
where \(I\) and \(U\) represent the intersection and union. For overlapping boxes (\( |x - \hat{x}| < w \)), the gradient with respect to \(\hat{x}\) is derived as:
\vspace{-1mm}
\begin{equation} 
\frac{\partial \mathcal{L}_{\text{IoU}}}{\partial \hat{x}} = \frac{2}{w^2 - d^2} \cdot \text{sign}(x - \hat{x}), 
\label{eq:iou_grad}
\end{equation}
the Hessian can be derived as:
\begin{equation} 
\frac{\partial^2 \mathcal{L}_{\text{IoU}}}{\partial \hat{x}^2} = \frac{4w}{(w^2 - d^2)^2},
\label{eq:iou_hessian}
\end{equation}
where the gradient scales inversely with $w$ and the Hessian with $w^3$. Consequently, smaller objects exhibit larger gradients and sharper loss curvature. This can lead to more significant fluctuations in the loss landscape and oscillations near the ground truth, resulting in instability during optimization.
Derivations of Eq.\,(\ref{eq:hessian_xy}), Eq.\,(\ref{eq:hessian_wh}), Eq.\,(\ref{eq:iou_grad}), and Eq.\,(\ref{eq:iou_hessian}) are provided in the supplementary material.

\subsection{Uncertainty-Aware Gradient Stabilization}
\label{sec:gradient_stability}
To address the gradient instability problem in small object localization, we propose the Uncertainty-Aware Gradient Stabilization (UGS) method, depicted in Fig.~\ref{architecture}. UGS integrates three components: a classification-based localization objective that generates bounded and confidence-driven gradients, an uncertainty minimization loss to further stabilize optimization, and an uncertainty-guided refinement module that leverages adversarial perturbations to identify and refine regions of high uncertainty.
\subsubsection{Classification-based Localization}
Building upon GFL V1~\cite{li2020generalized},  we propose a classification-based localization objective that quantizes continuous regression targets into interval non-uniform discrete grid representations. 
Specifically, the continuous regression range $[-\alpha, \alpha]$ for each target value $T$ is partitioned into $n+1$ uniformly spaced intervals, yielding the discrete grid set $Y = \{\mathbf{y}_0, \mathbf{y}_1, \dots, \mathbf{y}_n\}$. 
The ground truth $T$ is mapped to adjacent grids $i_l$ and $i_r$ using two-hot soft targets~\cite{li2020generalized}:  
\begin{equation}
\mathbf{p}_i^* =
\begin{cases}
|\mathbf{y}_i - T| \cdot \frac{n+1}{2\alpha}, & \text{if } i = i_l \text{ or } i_r, \\
0, & \text{otherwise.}
\end{cases}
\label{eq:pi}
\end{equation}
For small objects, the regression targets are distributed in a limited range.
When transformed into soft targets, the head range of the target distribution occupies few grids, leading to class imbalance and ineffective training.
Furthermore, in offset-based label systems~\cite{zhang2020bridging,ren2015faster}, target values decrease with training and are constantly assigned to the same grid interval after some iterations, further hindering optimization.
To address the issues, we design interval non-uniform (IN) labels through exponential grid spacing:
\begin{equation}
\mathbf{y}_i^\text{IN} = \text{sign}(\mathbf{y}_i) \cdot \frac{\alpha}{e^{\alpha\beta} - 1} \left(e^{\beta|\mathbf{y}_i|} - 1\right),
\label{eq:in_labels}
\end{equation}
where \(\mathbf{y}_{i}=-\alpha+\frac{2\alpha i}{n}\) are uniformly spaced grids within $[-\alpha, \alpha]$, and $\beta$ is a factor controlling the density of grid points. A larger $\beta$ results in denser grids near zero, balancing the localization targets for small objects and facilitating better optimization.

For optimization, the framework minimizes the cross-entropy loss between the predicted distribution $\mathbf{p}$ and the ground truth $\mathbf{p}^*$:
\begin{equation}
\mathcal{L}_\text{CE} = -\mathbf{p}_{i_l}^* \log \mathbf{p}_{i_l} - \mathbf{p}_{i_r}^* \log \mathbf{p}_{i_r}.
\label{eq:ce}
\end{equation}
where the continuous coordinate prediction can be restored as $\hat{T}=\sum_{i=0}^{n}\mathbf{p}_{i}*\mathbf{y}_{i}$.
We demonstrate that the gradient of the classification-based localization objective is both bounded and confidence-driven:
\begin{equation}
    \begin{aligned}
         \frac{\partial \mathcal{L}_\text{CE}}{\partial \mathbf{l}_{i}} 
        &= \mathbf{p}_{i} - \mathbf{p}_{i}^{*}, \\
        & = \begin{cases} 
        \mathbf{p}_{i} - |\mathbf{y}_{i}-T| \cdot \frac{n+1}{2\alpha}, & i=i_l, i_r \\
        \mathbf{p}_{i}, & \text{otherwise}, 
        \end{cases}
    \end{aligned}
\label{eq:ce_grad}
\end{equation}
which shows that the gradient magnitude of the cross-entropy loss with respect to logits $|\mathbf{p}_i - \mathbf{p}_i^*|$ is bounded within $[0,1]$ across object scales, mitigating instability for small objects. 
Additionally, the gradient magnitude is proportional to the confidence difference $|\mathbf{p}_i - \mathbf{p}_i^*|$ and enables confidence-driven learning, where large differences lead to more aggressive updates and small differences result in fine-grained refinements, facilitating precise localization.
 
%

%
\subsubsection{Uncertainty Minimization}
\label{sec:um}
To further stabilize training, we explicitly model and minimize the prediction uncertainty in the confidence distribution \(\mathbf{p}\):
\begin{equation}
\mathcal{L}_{\text{UM}} = \mathcal{H}(\mathbf{p}) = -\sum_{i=0}^{n} \mathbf{p}_i \log \mathbf{p}_i,
\label{eq:um}
\end{equation}
where \(\mathcal{H}(\mathbf{p})\) represents the entropy of the distribution \(\mathbf{p}\)~\cite{shannon1948mathematical}. 
By minimizing the entropy, the loss can suppress uncertain predictions that amplify gradient variance and reduce the coordinate prediction variance, stabilizing the optimization. 
Moreover, it penalizes high-confidence errors less aggressively than the cross-entropy objective through:
\begin{equation}
\frac{\partial \mathcal{L}_{\text{UM}}}{\partial \mathbf{l}_i} = -\left(\log \mathbf{p}_i + 1\right),
\label{eq:um_grad}
\end{equation}
which benefits small objects as they tend to generate noisy and uncertain predictions due to insufficient feature representations~\cite{li2022generalized}, exhibited in Fig.~\ref{fig:qualitative_case} (b).
The UM loss also maintains stable updates across object scales via bounded gradients.

\subsubsection{Uncertainty-guided Refinement}
\label{sec:methods-ur}
Inspired by prior works in beneficial noise learning~\cite{ishii2019training}, we propose an Uncertainty-guided Refinement (UR) module that leverages perturbations derived from the uncertainty minimization loss ($\mathcal{L}_{\text{UM}}$) to identify and refine regions of high uncertainty. This approach also enhances feature robustness and stabilizes the optimization process for small objects.

\begin{figure}[t]
\begin{center}
\centering
    \centerline{\includegraphics[width=\linewidth]{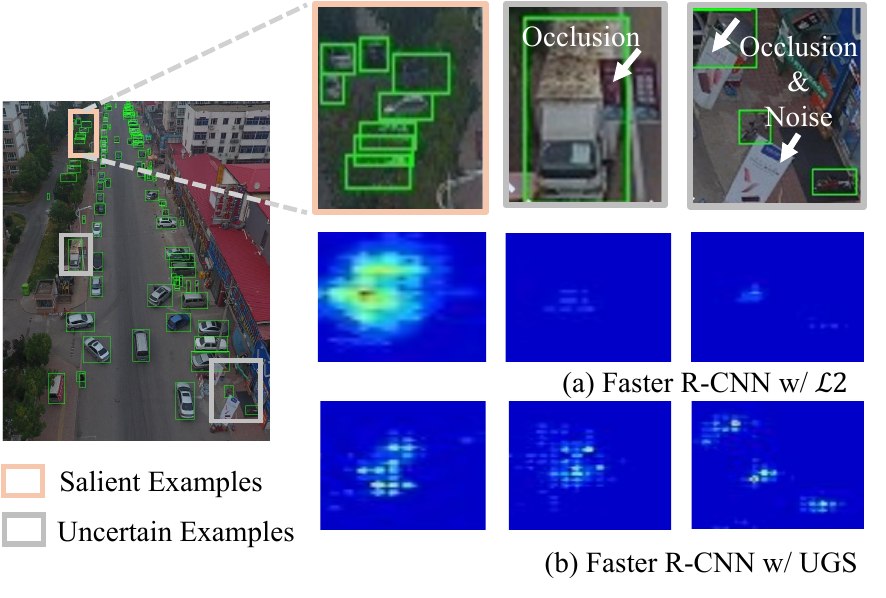}}
    \caption{Comparing the gradient magnitudes maps~\cite{guo2021distilling} using a 1$\times$ Faster R-CNN detector with ResNet-18 backbone. In (a), normal-scale objects converge by epoch 12 while the gradients for some small objects remain salient, indicating a convergence challenge. In (b), the gradients in the salient object are less pronounced. }
\label{results_gradients}
\vspace{-4mm}
\end{center}
\end{figure}

The UR formulation for FPN~\cite{lin2017feature} layer \( \mathbf{P}_i \) establishes a min-max objective:
\begin{equation}
\min_{\mathbf{P}_i, \theta_i} \left( \max_{\|\epsilon_{i}\| \leq \rho} \mathcal{L}_{\text{UM}}(\mathbf{P}_i + \epsilon_{i}) + \gamma \|\mathbf{P}_i\|_2^2 \right),
\label{eq:API-objective-um}
\end{equation}
where \(\mathcal{L}_{\text{UM}}\) quantifies prediction uncertainty via entropy minimization.
The inner optimization introduces adversarial perturbation \(\epsilon_{i}\) into the feature space of \(\mathbf{P}_i\), with \(\rho\) controlling the perturbation magnitude and \(\gamma\) regulating the regularization strength. Here, \(\theta_i\) represents the model parameters at the \(i\)-th layer. 
Following~\cite{foret2020sharpness}, we derive the closed-form perturbation under \(L_2\)-norm:
\begin{equation}
\epsilon_{i}^* \approx \rho \cdot \frac{\nabla_{\mathbf{P}_i} \mathcal{L}_{\text{UM}}(\mathbf{P}_i)}{\|\nabla_{\mathbf{P}_i} \mathcal{L}_{\text{UM}}(\mathbf{P}_i)\|_2},
\label{eq:API-epsilon-approx-um-l2}
\end{equation}
where the perturbation \(\epsilon_{i}^*\) targets regions where \(\mathcal{L}_{\text{UM}}\) exhibits high sensitivity to activation changes, indicating areas of high uncertainty.
The introduced perturbation can amplify the refinement~\cite{kim2023feature} of these uncertain regions while preserving stable updates in regions with high confidence.
Fig.~\ref{results_gradients} shows that the method can learn about occluded objects and noise that resembles objects.
Additionally, the adversarial objective encourages the learning of robust feature representations~\cite{hu2022adversarial}, further stabilizing the training process for small objects. 

We formulate the overall training objective as:
\begin{equation}
\mathcal{L}_\text{localization} = \mathcal{L}_\text{CE} + \lambda \mathcal{L}_{\text{UM}} + \underbrace{\gamma \cdot \sum_{i=1}^{N} \mathcal{L}_{i}^{\text{ur}}(\mathbf{P}_i + \epsilon_i^*)}_{\text{Uncertainty-Guided Refinement Loss}},
\label{eq:aux}
\end{equation}
where \(\lambda,\gamma\) are balancing hyperparameters, and \(\mathcal{L}_{i}^{\text{ur}}\) denotes the layer-wise uncertainty-guided refinement loss. 
Described in Sec.~\ref{fig:gradient_analysis}, our approach reduces gradient variance by 2.5$\times$ compared to Smooth-$\mathcal{L}_1$ loss for small objects.
\begin{table*}[t!]  
\centering  
\small
\caption{Detection performance on the VisDrone {\tt val} set~\cite{zhu2018visdrone}. $+(\cdot)$ indicates the improvement over the baseline, \textbf{bold} denotes the best results. Baselines include general and small-object-oriented detectors. $^\ast$ denotes re-implementation results.}  
\label{det1}  
\begin{tabularx}{0.8\textwidth}{l|c|XX|XXX}  
\hline
{Method}  & {Year} & {AP} & {AP$_{50}$} & {AP$_{s}$} & {AP$_{m}$} & {AP$_{l}$}\\  
\hline
QueryDet~\cite{yang2022querydet} & 2022 & 28.3 & 48.1 & - & -& -  \\ 
DetectoRS~\cite{qiao2021detectors}  & 2021 & 29.4 & 49.3 & - & - & -  \\  
CZDet~\cite{meethal2023cascaded}  & 2023 & 33.2 & 58.3 & - & -& -  \\  
DQ-DETR~\cite{huang2024dq} & 2024 & 37.0 & 60.9  & - & - & -  \\
\hline
FCOS~\cite{tian2019fcos} & 2019 & 19.9 & 37.7 & 11.4 & 30.7 & 35.2 \\
\textbf{w/ UGS} & - & \textbf{22.4} & \textbf{39.7} & \textbf{13.0} & \textbf{33.2} & \textbf{37.8} \\
\hline
Faster R-CNN~\cite{ren2015faster} & 2015 & 21.3 & 36.4 & 12.8 & 32.9 & 38.9 \\
\textbf{w/ UGS} & - & \textbf{24.2} & \textbf{41.3} & \textbf{15.8} & \textbf{37.6} & \textbf{42.5} \\
\hline
Cascade R-CNN~\cite{cai2018cascade} & 2018 & 22.4 & 37.0 & 13.1 & 34.1 & 42.1 \\
\textbf{w/ UGS} & - & \textbf{24.4} & \textbf{40.1} & \textbf{15.8} & \textbf{37.8} & \textbf{44.3} \\
\hline
GFL V1~\cite{li2020generalized} & 2020 & 28.4 & 50.0 & 15.9 & 39.7 & 42.6 \\
w/ CEASC~\cite{du2023adaptive} & 2023 & 28.7 & 50.7 & - & - & - \\
\textbf{w/ UGS} & - & \textbf{31.2} & \textbf{53.0} & \textbf{19.2} & \textbf{42.6} & \textbf{54.4} \\
\hline
DINO-5scale~\cite{zhang2022dino} & 2023 & 35.5 & 58.0 & 22.4 & 45.6	& 51.2 \\
\textbf{w/ UGS} & - & \textbf{38.1} & \textbf{61.9} & \textbf{24.2} & \textbf{47.3} & \textbf{56.6} \\ 
\hline
\end{tabularx}  
\end{table*}  

\section{Experiments}
\subsection{Datasets and Implementation Details}
\subsubsection{Datasets} 
We evaluate the proposed method on \textbf{VisDrone}~\cite{zhu2018visdrone}, \textbf{SODA-A}\cite{cheng2023towards}, \textbf{COCO {\tt 2017}}~\cite{lin2014microsoft}, and the \textbf{PASCAL VOC} dataset ~\cite{Everingham10}.
The main experiments were conducted using the {VisDrone}~\cite{zhu2018visdrone} dataset, which contains a high proportion of small object instances. The dataset consists of 10,209 drone-shot images divided into a training set (6,471 images), a validation set (548 images), and a test set (3,190 images). Meanwhile, we conduct experiments on {SODA-A}~\cite{cheng2023towards} dataset, which comprises 2513 high-resolution images of aerial scenes, which has 872069 instances annotated with \textit{oriented} rectangle box annotations over 9 classes. 
We also perform experiments on two general object detection benchmarks: the {COCO {\tt 2017}}~\cite{lin2014microsoft} dataset and the {PASCAL VOC}~\cite{Everingham10} dataset.
\begin{table}[t]
\centering
\small
\caption{Detection performance of YOLO-based detectors on the VisDrone {\tt val}~\cite{zhu2018visdrone}, using CSP-D53~\cite{wang2021scaled} as backbone. $+(\cdot)$ indicates the performance improvement, and \textbf{bold} denotes the best results.}
\begin{tabularx}{0.48\textwidth}{l|c|XXX}
\hline
 Method & Input Size & AP & AP$_{s}$ & AP$_{m}$ \\  
\hline
TPH-YOLOv5-x~\cite{zhu2021tph}  &   \multirow{2}{*}{(640, 640)}      & 26.5     & 14.1            & 22.0  \\ 
\textbf{w/ UGS}  &    & \textbf{29.0} & \textbf{16.7} & \textbf{23.9} \\ 
\hline
TPH-YOLOv5-x~\cite{zhu2021tph} & \multirow{2}{*}{(1536, 1536)} & 39.2   & 22.4  & 34.7 \\
\textbf{w/ UGS}  &   & \textbf{41.7} & \textbf{24.2} & \textbf{36.2} \\ 
\hline
\end{tabularx}
\label{det2}
\end{table}
\subsubsection{Implementation Details.} 
We conduct the experiments on a computer with 4 NVIDIA RTX 3090 GPUs. 
The models are implemented using PyTorch~\cite{paszke2017automatic}, with core functionalities built upon the MMDetection framework~\cite{chen2019mmdetection}.
For the VisDrone dataset~\cite{zhu2018visdrone}, we use an input resolution of 1333$\times$800 with a batch size of 16.
For fair comparisons, we utilize default optimization parameters and 1$\times$ learning schedule without multi-scale training. 
We utilize the SGD optimizer with initial learning rate of 2e-2, decayed by 0.1 at epoch 8 and 11, and weight decay of 1e-4. 
We evaluate our method on two state-of-the-art small object detectors. For YOLO-based TPH-YOLOv5~\cite{zhu2021tph}, we train the model for 80 epochs using the default hyperparameters from~\cite{zhu2021tph}. We evaluate under two input resolutions—640$\times$640 and 1536$\times$1536, with batch sizes of 16 and 4, respectively.
%
We also train a transformer-based detector, DINO~\cite{zhang2022dino}, with 5-scale feature maps for 24 epochs as a baseline.
Training employs the AdamW optimizer with initial learning rate 2e-4, and adopts DETR-style augmentation~\cite{zhudeformable} including random cropping and multi-scale resizing.
At test time, the maximum number of predictions per image is increased to 1500 following DQ-DETR~\cite{huang2024dq} to accommodate dense small-object scenarios.

\begin{table*}[t!]
\centering
\small
\caption{Main results with various frameworks and FPN features on SODA-A. Note that models are trained on the SODA-A {\tt train} and validated on the SODA-A {\tt test}. We report APs (\%) with different IoU threshold and APs (\%) for objects in various sizes based on the SODA-A criterion. The \textbf{bold} denotes the best result.}
\begin{tabularx}{0.9\textwidth}{l|XXX|XXXX}
\hline
{Method} & {AP}    & {AP$_{0.5}$}    & {AP$_{0.75}$}   & {AP$_{ eS}$}   & {AP$_{rS}$}    & {AP$_{gS}$}   & { AP$_{ N}$ }    \\ \hline
Rotated RetinaNet ~\cite{lin2017focal} & 22.3          & 57.7          & 10.6          & 7.7          & 18.0          & 28.0          & 23.0         \\
\textbf{w/ UGS}   & \textbf{26.8} & \textbf{63.2} & \textbf{15.9} & \textbf{8.9} & \textbf{21.7} & \textbf{34.9} & \textbf{27.5} \\   \hline
Rotated FCOS ~\cite{tian2019fcos}  & 32.6          & 69.4          & 24.8          & 11.5         & 30.0          & 42.7          & 41.6         \\
\textbf{w/ UGS}   & \textbf{35.2} & \textbf{72.9} & \textbf{28.7} & \textbf{12.3} & \textbf{30.5} & \textbf{46.3} & \textbf{39.2} \\  \hline
Oriented Reppoints~\cite{li2022oriented} & 25.4          & 60.0          & 16.8          & 8.6          & 21.9          & 29.6          & 27.8         \\ 
\textbf{w/ UGS} & \textbf{28.2} & \textbf{64.5} & \textbf{19.7} & \textbf{9.8} & \textbf{23.8} & \textbf{33.6} & \textbf{31.2} \\ \hline
Oriented RCNN ~\cite{li2022oriented} & 34.4          & 70.7          & 28.6          & 12.5          & 28.6          & 44.5          & 36.7         \\
\textbf{w/ UGS}  & \textbf{36.0} & \textbf{73.1} & \textbf{30.3} & \textbf{13.7} & \textbf{30.2} & \textbf{47.8} & \textbf{38.1} \\  \hline
\end{tabularx}
\label{SODA-A}
\end{table*}
\subsection{Experiments on VisDrone}
We evaluate the proposed UGS method with various baselines and two state-of-the-art small object detectors on the VisDrone {\tt val} set~\cite{zhu2018visdrone}.  
As shown in Tab.\,\ref{det1}, UGS enhances all baseline detectors by approximately 2\% in AP, a notable improvement. 
Specifically, UGS boosts the performance of the anchor-free FCOS~\cite{tian2019fcos} by 2.5\% in AP and 1.6\% in AP$_{s}$.
Integrated with two-stage detectors, UGS improves Faster R-CNN~\cite{ren2015faster} by 2.6\% AP and 1.5\% AP$_{s}$, while enhancing Cascade R-CNN~\cite{cai2018cascade} by 2.0\% AP and 2.7\% AP$_{s}$, yielding consistent performance gains.
Furthermore, we apply UGS to GFL V1~\cite{li2020generalized}, a strong general object detector that employs a classification-based localization objective. UGS improves GFL V1 by a significant 3.5\% AP, surpassing the state-of-the-art comparison CEASC~\cite{du2023adaptive}.
%
%
In addition, we evaluate UGS's compatibility with DETR-based architecture DINO~\cite{zhang2022dino}. UGS boosts the performance of the DINO-5scale baseline by 2.6\% AP, surpassing the prior art DQ-DETR~\cite{huang2024dq}.
Tab.~\ref{det2} highlights the compatibility of UGS with YOLO-based architectures. Specifically, UGS improves TPH-YOLOv5 by 2.5\% AP at both $640^2$ and $1536^2$ resolutions. In particular, UGS achieves 41.7\% AP in the VisDrone validation set without test-time augmentation.
\begin{table}[t]
\small
\centering
\caption{Experiments with Faster R-CNN on VOC~\cite{Everingham10} and COCO~\cite{lin2014microsoft}. $+(\cdot)$ indicates the performance improvement, and \textbf{bold} denotes the best results. ‘-’ indicates that the result is not reported or not publicly available.}
\begin{tabularx}{\linewidth}{l|l|X|ll}
\hline
{Backbone} &{Method}  & {Dataset} & {AP}   & {AP$_{s}$}  \\ 
\hline
\multirow{4}{*}{R-50}                                & Faster R-CNN                   & \multirow{2}{*}{VOC}  & 78.4     &    -       \\
& \textbf{w/ UGS}   &   & \textbf{82.2}  & - \\ \cline{2-5} 
& Faster R-CNN  & \multirow{2}{*}{COCO} &  37.4  &  21.2 \\
& \textbf{w/ UGS} &   & \textbf{38.9}  & \textbf{22.6} \\ 
\hline
\multirow{4}{*}{R-101}  & Faster R-CNN   & \multirow{2}{*}{VOC}  & 79.9 &     -      \\
& \textbf{w/ UGS}   &  & \textbf{81.7}  & - \\ \cline{2-5} 
& Faster R-CNN    & \multirow{2}{*}{COCO} &   39.4   &    22.4   \\
& \textbf{w/ UGS}   &  & \textbf{41.4}  & \textbf{24.9}  \\ 
\hline
\end{tabularx}
\label{voc}
\end{table}
\subsection{Experiments on SODA-A}
We further validate the UGS method on the aerial rotated small object detection dataset SODA-A~\cite{cheng2023towards}.
As shown in Tab.\,~\ref{SODA-A}, the proposed UGS method effectively improves the performance of both anchor-free and anchor-based rotating detectors. 
Specifically, for the single-stage anchor-based model, Rotated RetinaNet~\cite{lin2017focal}, UGS improved the AP by 4.5\%. 
For the anchor-free model, Rotated FCOS, UGS enhanced the AP by 2.6\%. 
For the proposal-based Oriented RCNN, UGS boosted the AP by 1.6\%.
%
%

%
\subsection{Experiments on COCO and VOC}  
To evaluate the generalizability of UGS, we conduct experiments on two general object detection benchmarks: PASCAL VOC~\cite{Everingham10} and COCO~\cite{lin2014microsoft}.
As shown in Tab.~\ref{voc}, UGS achieves significant improvements over the baselines on VOC. Under the ResNet-50 backbone, UGS improves Faster R-CNN  by 3.8\% AP, raising the performance from 78.4\% to 82.2\%. With the ResNet-101 backbone, UGS achieves a 1.8\% improvement.
On the COCO dataset, UGS demonstrates consistent improvements, particularly for small objects (AP$_{s}$). With the ResNet-50 backbone, UGS improves overall AP by 1.5\% and AP$_{s}$ by 1.4\%. For ResNet-101, the improvements are more pronounced. These results underscore UGS’s effectiveness in improving small object detection while maintaining strong performance across all object sizes.  
\subsection{Training Cost of UGS} 
We analyze the training cost of our UGS on 1 NVIDIA RTX 3090 GPU with $24$ GB memory. 
We test them using a batch size of $2$ and input resolution of $1333\times800$ at each iteration. 
As shown in Tab.~\ref{time}, UGS introduces 15\%,  0.6\%, 13\% increase in training time, computational and memory cost, which is moderate.
\subsection{Gradient Variance Analysis}
\label{sec:gradient_analysis}
To quantify gradient stability, we measure the gradient variance during training for both our method (UGS) and the Smooth-$\mathcal{L}_1$ baseline. For small objects ($w_a < 32$), we compute the variance of localization loss gradients across 100 training iterations and report the mean variance ratio, defined as the ratio of the gradient variance of Smooth-$\mathcal{L}_1$ to that of UGS.
\begin{figure}[t]
    \centering
    \includegraphics[width=\linewidth]{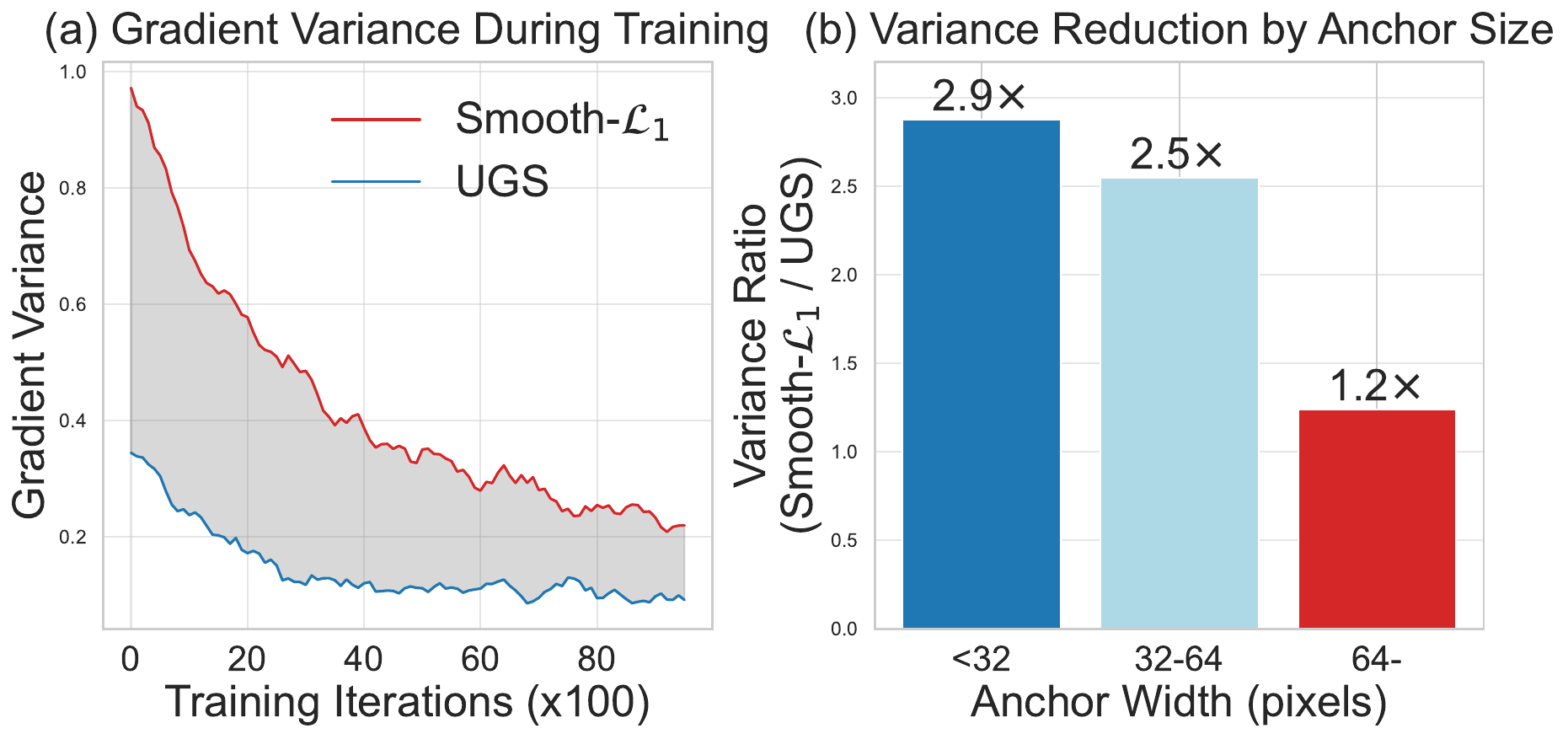}
    \caption{Gradient variance analysis for small objects. (a) Variance of localization loss gradients across training iterations. (b) Mean variance ratio (UGS vs. Smooth-$\mathcal{L}_1$) grouped by anchor size. Our method reduces variance by 2.9$\times$ for $w_a < 32$ pixels.}
    \label{fig:gradient_analysis}
\end{figure}
\begin{table}[t!]
\centering
\small
\caption{Comparing the training cost of Baseline and our UGS using various detection frameworks using ResNet-50 backbone.}
\begin{tabularx}{0.48\textwidth}{l|c|c|c|c}
\hline
 Method & AP  & Time $_{({\rm s} / {\rm batch})}$ & GFLOPs & Params (M) \\ \hline
 FCOS &  19.9  & 0.151  & 127.25 & 42.40   \\
w/ UGS  & 22.4    &   0.174    & 128.06 & 48.12      \\ \hline
\end{tabularx}
\vspace{-2mm}
\label{time}
\end{table}
Results in Fig.~\ref{fig:gradient_analysis} demonstrate that UGS achieves a 2.9$\times$ lower gradient variance compared to Smooth-$\mathcal{L}_1$ for small objects. This validates our analysis in Sec.~\ref{sec:gradient_stability} that classification-based localization objective and the proposed uncertainty-aware designs lead to smoother optimization landscapes and more stable gradients. 
\subsection{Visualizations}
We show in Fig.~\ref{results_gradients} that our design leads to a more stable training process, where the gradient magnitude of salient small object examples becomes less pronounced compared to the $\mathcal{L}_2$ loss.
Fig.~\ref{results_gradients} also highlights UGS’s ability to learn from uncertain examples, including occluded objects, objects in cluttered backgrounds, and noise-resembling objects. 
\begin{table}[t!]
\small
\centering
\caption{Ablation study of UGS components on VisDrone validation set using Faster R-CNN. Improvements over $\mathcal{L}_2$ baseline are marked in blue. UR denotes the uncertainty-guided refinement loss.}
\begin{tabularx}{0.48\textwidth}{@{}lXXXcccc@{}}
\hline
{Method} & $\lambda$ & $\gamma$ & $\rho$ & {IN} & {AP} & {AP$_{50}$} & {AP$_{s}$} \\ 
\hline
$\mathcal{L}_2$ (Baseline) & - & - & - & - & 21.3 & 36.4 & 12.8 \\
Smooth-$\mathcal{L}_1$ & - & - & - & - & 21.5 & 36.4 & 12.7 \\
IoU-loss~\cite{yu2016unitbox} & - & - & - & - & 21.8 & 37.0 & 13.0  \\
Gradient Clipping  & - & - & - & - & 21.6 & 36.7 & 12.9 \\
Bayesian YOLO~\cite{kraus2019uncertainty} & - & - & - & - & 22.0 & 37.3 & 13.1 \\
\hline
\multicolumn{8}{@{}l}{\textit{Classification-based Localization}} \\
$\mathcal{L}_\text{CE}$ & - & - & - & - & 22.1 & 37.1 & 13.2 \\
$\mathcal{L}_\text{CE}$+IN & - & - & - & ✓ & 22.5 & 38.2 & 13.4 \\
\hline
\multicolumn{6}{@{}l}{\textit{CE + Uncertainty Minimization (UM)}} \\
$\mathcal{L}_\text{CE} + \lambda\mathcal{L}_\text{UM}$ & 0.1 & -  & - & ✓ & 22.7 & 38.2 & 13.6 \\
$\mathcal{L}_\text{CE} + \lambda\mathcal{L}_\text{UM}$ & 0.5 & - & - & ✓ & 22.9 & 38.4 & 13.6 \\
$\mathcal{L}_\text{CE} + \lambda\mathcal{L}_\text{UM}$ & 1.0 & - & - & ✓ & 22.6 & 37.2 & 13.7 \\
\hline
\multicolumn{6}{@{}l}{\textit{CE + UM + Uncertainty-guided Refinement (UR)}} \\
$\mathcal{L}_\text{CE} + \lambda\mathcal{L}_\text{UM} + \gamma\mathcal{L}^\text{ur}$ & 0.5 & 0.1 & 0.5 & ✓ & 23.5 & 39.0 & 14.1 \\
$\mathcal{L}_\text{CEf} + \lambda\mathcal{L}_\text{UM} + \gamma\mathcal{L}^\text{ur}$ & 0.5 & 0.5 &  0.5 & ✓ & \textbf{24.2} & \textbf{41.3} & \textbf{15.8} \\
$\mathcal{L}_\text{CE} + \lambda\mathcal{L}_\text{UM} + \gamma\mathcal{L}^\text{ur}$ & 0.5 & 0.5 &  1 & ✓ & 23.2 & 38.7 & 13.9 \\
\hline
\end{tabularx}
\label{ablation}
\vspace{-4mm}
\end{table}
%
\vspace{-1mm}
\subsection{Ablation Study}
In the following experiments, we demonstrate that UGS leads to consistent performance increment when applied on baseline detectors and explore the best-performing structure by ablating and tuning each component. We use Faster R-CNN~\cite{ren2015faster} with ResNet-50~\cite{he2016deep} as the baseline and conduct all tests on the VisDrone {\tt val} dataset~\cite{zhu2018visdrone}.

\noindent{\bf{Effectiveness of Components.}}
According to Tab.~\ref{ablation}, replacing $\mathcal{L}_2$ with the classification-based localization objective ($\mathcal{L}_\text{CE}$) significantly improves performance, achieving 22.1 AP.
Utilizing IN labels further increases performance by 1.2\% AP and 0.6\% AP$_{s}$, demonstrating the importance of non-uniform quantization for ensuring balanced supervision for small object detection. 
Adding uncertainty minimization loss with $\lambda = 0.5$ increases performance to 22.9 AP, validating the effectiveness of entropy-based uncertainty modeling.
The full UGS framework, incorporating CE, UM, and the uncertainty-guided refinement module ($\mathcal{L}^\text{ur}$), achieves the best performance with 24.2 AP, 41.3 AP$_{50}$ and 15.8 AP$_{s}$. This represents a 2.9 AP improvement over the $\mathcal{L}_2$ baseline, which is significant. 
%
We also evaluate existing gradient stabilization techniques for comparison, including normalized regression losses, gradient clipping ($\| \nabla \| \leq 1.0$), and uncertainty-aware method Bayesian YOLO~\cite{kraus2019uncertainty}.
While the methods effect in mitigating gradient instability, their improvements are relatively marginal compared to UGS.

\noindent{\bf Effect of \(\alpha\), \(\beta\), and \(n\) in Interval Non-uniform (IN) Label Quantization.} 
We investigate the impact of three key parameters in UGS’s interval non-uniform (IN) label quantization: the range parameter \(\alpha\), the IN modulator \(\beta\), and the grid number \(n\).  
Experiments show that a moderate range of \(\alpha = 2\) for RPN and \(\alpha = 5\) for R-CNN yields the best performance improvements, achieving 1.2\% AP and 0.6\% AP$_{s}$. Notably, the performance remains robust across different values of \(\alpha\), indicating that UGS is not overly sensitive to this parameter.  
A smaller \(\beta\) (e.g., 1.0) generally leads to better performance, as it creates a denser grid distribution around zero. This design is particularly effective for capturing finer details in small object instances, where precise localization is critical.  
UGS demonstrates flexibility in the choice of grid number \(n\), with performance remaining stable across different configurations. 

\begin{table}[t!]
\small
\centering
\caption{Parameter analysis of UGS components on Faster R-CNN: Effects of range ($\alpha$), IN modulator ($\beta$), and grid number ($n$). Best results per stage are in bold.}
\begin{tabularx}{0.48\textwidth}{@{}cccccclll@{}}
\hline
{Stage} & {Parameter} & $\alpha$ & $\beta$ & $n$ & {AP} & {AP$_{50}$} & {AP$_{s}$} \\ 
\hline
\multirow{6}{*}{RPN} & \multirow{4}{*}{$\alpha$-$\beta$} & 2 & 1.0 & 10 & \textbf{21.9} & \textbf{37.4} & \textbf{13.0} \\
 & & 2 & 1.5 & 10 & 21.8 & 37.2 & 13.1 \\ \cmidrule(lr){3-8}
 & & 3 & 1.0 & 10 & 21.8 & 37.2 & 13.1 \\
 & & 3 & 1.5 & 10 & 21.7 & 37.1 & 13.0 \\ \cmidrule(lr){2-8}
 & \multirow{4}{*}{Grid Number} & \multirow{4}{*}{2} & \multirow{4}{*}{1.0} & 2 & 21.8 & 36.9 & 12.9 \\
 & & & & 4 & 21.9 & 37.1 & 13.0 \\
 & & & & 10 & \textbf{21.9} & \textbf{37.4} & \textbf{13.0} \\
 & & & & 20 & 21.8 & 37.1 & 13.0 \\ 
\hline
\multirow{6}{*}{R-CNN} & \multirow{4}{*}{$\alpha$-$\beta$} & 4 & 1.0 & 5 & 22.2 & 37.6 & 13.1 \\
 & & 4 & 1.5 & 5 & 21.9 & 37.2 & 13.0 \\ \cmidrule(lr){3-8}
 & & 5 & 1.0 & 5 & \textbf{22.5} & \textbf{38.2} & \textbf{13.4} \\
 & & 5 & 1.5 & 5 & 22.0 & 37.0 & 13.1 \\ \cmidrule(lr){2-8}
 & \multirow{3}{*}{Grid Number} & \multirow{3}{*}{5} & \multirow{3}{*}{1.0} & 5 & \textbf{22.5} & \textbf{38.2} & \textbf{13.4} \\
 & & & & 10 & 22.3 & 37.6 & 13.2 \\
 & & & & 20 & 22.4 & 37.9 & 13.4 \\ 
\hline
\end{tabularx}
\label{merged_params}
\vspace{-4mm}
\end{table}

%
\section{Conclusions}
In this paper, we demonstrate that conventional object localization methods tend to produce unstable gradients on small objects and introduce an Uncertainty-Aware Gradient Stabilization (UGS) method to rectify gradients. 
UGS quantizes continuous labels into interval non-uniform discrete representations as supervision. 
In optimization, UGS employs a classification-based localization objective that generates bounded and confidence-driven gradients. Further, UGS integrates a dual uncertainty-aware mechanism to enhance the robustness of localization. 
UGS consistently improves baselines and state-of-the-art small object detectors, demonstrating compatibility across different detection architectures.

\noindent\textbf{Acknowledgments.}  
This work was supported by the National Key Research and Development Program of China (No. 2023YFC3306401).
This research was also supported by the National Natural Science Foundation of China (No. 61827901, 623B2016, 62406298),
the Zhejiang Provincial Natural Science Foundation (No. LD24F020007),
the Beijing Natural Science Foundation (No. L223024, L244043, Z241100001324017),
the “One Thousand Plan” projects in Jiangxi Province (No. Jxsq2023102268),
and the Fundamental Research Funds for the Central Universities (CUC25QT17).
{
    \small
    \bibliographystyle{ieeenat_fullname}
    \bibliography{main}
}
\end{document}